\documentclass[11pt,a4paper]{article}
\usepackage[hyperref]{acl2018}
\usepackage{times}
\usepackage{latexsym}

\usepackage{graphicx}
\usepackage{url}

\aclfinalcopy 

\title{Classifying Patent Applications with Ensemble Methods}

\author{Fernando Benites\textsuperscript{1}, Shervin Malmasi\textsuperscript{2,3}, Marcos Zampieri\textsuperscript{4} \\
  \textsuperscript{1}Zurich University of Applied Sciences, Switzerland \\
  \textsuperscript{2}Harvard Medical School, United States \\
    \textsuperscript{3}Macquarie University, Australia \\
  \textsuperscript{4}University of Wolverhampton, United Kingdom \\
  {\tt benf@zhaw.ch, shervin.malmasi@mq.edu.au, m.zampieri@wlv.ac.uk} \\}

\date{}

\begin{document}
\maketitle
\begin{abstract}
\vspace{-1mm}
  We present methods for the automatic classification of patent applications using an annotated dataset provided by the organizers of the ALTA 2018 shared task - Classifying Patent Applications. The goal of the task is to use computational methods to categorize patent applications according to a coarse-grained taxonomy of eight classes based on the International Patent Classification (IPC). We tested a variety of approaches for this task and the best results, 0.778 micro-averaged F1-Score, were achieved by SVM ensembles using a combination of words and characters as features. Our team, BMZ, was ranked first among 14 teams in the competition.
\end{abstract}

\section{Introduction}

According to statistics of the World Intellectual Property Organization (WIPO),\footnote{\url{http://www.wipo.int/ipstats/en/}} the number of patent applications filled across the world keeps growing every year. To cope with the large volume of applications, companies and organizations have been investing in the development of software to process, store, and categorize patent applications with minimum human intervention.

An important part of patent application forms is, of course, composed of text. This has led to the widespread use of NLP methods in patent application processing systems as evidenced in Section \ref{sec:related}. One such example is the use of text classification methods to categorize patent applications according to standardized taxonomies such as the International Patent Classification (IPC)\footnote{\url{http://www.wipo.int/classifications/ipc/en/}} as discussed in the studies by \newcite{benzineb2011automated,fall2003automated}.

In this paper, we present a system to automatically categorize patent applications from Australia according to the top sections of the IPC taxonomy using a dataset provided by the organizers of the ALTA 2018 shared task on Classifying Patent Applications \cite{ALTA2018}.\footnote{\url{http://www.alta.asn.au/events/sharedtask2018/}} The dataset and the taxonomy are presented in more detail in Section \ref{sec:data}. Building on our previous work \cite{malmasi2016ltg,malmasi-zampieri:2017:VarDial1}, our system is based on SVM ensembles and it achieved the highest performance of the competition.

\section{Related Work}
\label{sec:related}

There have been a number of studies applying NLP and Information Retrieval (IR) methods to patent applications specifically, and to legal texts in general, published in the last few years.

Applications of NLP and IR to legal texts include the use of text summarization methods \cite{farzindar2004legal} to summarize legal documents and most recently, court ruling prediction. A few papers have been published on this topic, such as the one by \newcite{Katz14} which reported 70\% accuracy in predicting decisions of the US Supreme Court, \newcite{aletras2016predicting,medvedeva2018judicial} which explored computational methods to predict decisions of the European Court of Human Rights (ECRH), and \cite{sulea2017exploring,sulea2017predicting} on predicting the decisions of the French Supreme Court. In addition to the aforementioned studies, one recent shared task has been organized on court rule prediction \cite{zhong2018overview}.

 
Regarding the classification of patent applications, the task described in this paper, a related dataset WIPO-alpha was used in the experiments and it is often used in such studies.
The  WIPO-alpha consists of a different number of patents (in the thousands, but it grows every year) and is usually used in its hierarchical call form \cite{tikk2003experiment}. Recently, word embeddings and LSTMs were applied to the task \cite{8260665}. There, the experiments were hierarchically conducted but in a superficial manner.

Hoffmann et al. investigated in depth the hierarchical problem of WIPO-alpha with SVMs \cite{hofmann2003learning,tsochantaridis2004support,cai2007exploiting}. They showed that using a hierarchical approach produced better results. Many studies showed that evaluating a hierarchical classification task is not trivial and many measures can integrate the class ontology. Still, using multiple hierarchical measures can introduce bias \cite{BruckerBS11}. Yet, there was much improvement in the last 3-4 years in the text classification field. This is one reason, why, when reengaging again in the WIPO-alpha dataset, investigating only the top nodes of WIPO class ontology might be a good start for future successive tasks.

Finally, at the intersection between patent applications and legal texts in general, \newcite{wongchaisuwat2016} presented experiments on predicting patent litigation and time to litigation.

\section{Data}

\label{sec:data}

The dataset released by the organizers of the ALTA 2018 shared task consists of a collection of Australian patent applications. The dataset contains 5,000 documents released for training and 1,000 documents for testing. The classes relevant for the task consisted of eight different main branches of the WIPO class ontology as follows:

\begin{itemize}
\setlength\itemsep{0.1em}
\item A: Human necessities;
\item B: Performing operations, transporting;
\item C: Chemistry, metallurgy;
\item D: Textiles, paper;
\item E: Fixed constructions;
\item F: Mechanical engineering, lighting, heating, weapons, blasting;
\item G: Physics;
\item H: Electricity.
\end{itemize}

\noindent The documents were created using automated OCR and therefore, not thoroughly cleaned before release. For example, there were documents with expressions such as ``NA$\setminus\setminus$nparse failure'' and page numbers in the middle of paragraphs which made processing more challenging. We enhanced the dataset with  data from the WIPO-alpha repository gathered in October 2018 consisting of 46,319 training documents and 28,924 test documents. 
We also took a random sub-sample of 100,000 documents from the WIPO-en gamma English dataset, which contains 1.1 million patent documents in total.

We utilized all of the available text fields in the texts and concatenated them into a single document.

\section{Methodology}
\subsection{Preprocessing}

The documents come from different sources and authors, therefore no standard representation exists and there is high variation in formatting across the documents. Since we do not utilize document structure in our approach, we decided to eliminate it by collapsing the documents into a single block of text. This was done be replacing all consecutive non-alphanumeric characters with a single space. Next, we converted the text to lowercase and removed any tokens representing numbers.

\begin{table*}[tp]
\centering
\begin{tabular}{|l|c|c|c|}
\hline
            &\bf Training & \bf Public (Validation) & \bf Private (Test) \\
\hline
(1) Baseline 20k feats.& 0.709 & 0.710 & 0.692 \\
(2) Baseline 40k feats.& 0.715 & - & - \\
(3) Baseline w/ WIPO-alpha & 0.775 &0.758&0.744\\
(4) Semi-supervised  & 0.734 & 0.728 & 0.704 \\
(5) Ensemble w/ WIPO-alpha + gamma & 0.787 &0.776&0.778\\
\hline
\end{tabular}
\label{tab:resultsleaderboard}
\caption{F1-micro performance of the systems in training (10-fold CV), in the validation and in the test sets (train, public and  private leaderboard).}
\end{table*}

\subsection{Features}

For feature extraction we used and extended the methods reported in \newcite{malmasi-zampieri:2017:VarDial1}. Term Frequency (TF) of $n$-grams with $n$ ranging from 3 to 6 for characters and 1-2 for words have been used. Along with term frequency we calculated the inverse document frequency (TF-IDF) \cite{gebre2013} which resulted in the best single feature set for prediction.

\subsection{Classifier}

We used an ensemble-based classifier for this task. Our base classifiers are linear Support Vector Machines (SVM). SVMs have proven to deliver very good performance in a number of text classification problems. It was previously used for complex word identification \cite{malmasi2016ltg}, triage of forum posts \cite{malmasi2016predicting}, dialect identification \cite{malmasi-zampieri:2017:VarDial1}, hate speech detection \cite{malmasi2018challenges}, and court ruling prediction \cite{sulea2017exploring}.

\subsection{Systems}

We developed a number of different systems.
As baselines we employed single SVM models with TF-IDF, using the top 20k and 40k more frequent words as features, resulting in two models. We created a third baseline which included the WIPO-alpha data for training.

For system 4, we augmented system 3 with a semi-supervised learning approach similar to the submission by \newcite{jauhiainen2018heli} to the dialect identification tasks at the VarDial workshop \cite{zampieri2018language}. This approach consists of classifying the unlabelled test set with a model based on the training data, then selecting the predictions with the highest confidence and using them as new additional training samples. This approach can be very useful if there are few training samples and out-of-domain data is expected.

Finally, for system 5, we extended system 4 to be an ensemble of both word- and character-based models, and to include additional training data from the WIPO-alpha and WIPO-en gamma datasets, as described in \ref{sec:data}.

\section{Results}

In this section, we investigate the impact of the different systems and data. We give special attention to the competition results showing these in different settings. This is particularly interesting since the amount of data with WIPO-alpha and the vocabulary of the ALTA data without pre-processing was relatively large.

\subsection{Official Results}

We present the results obtained in the training stage, the public leaderboard, and the private leaderboard in Table \ref{tab:resultsleaderboard}. The shared task was organized using Kaggle\footnote{\url{https://www.kaggle.com/}}, a data science platform, in which the terms Public Leaderboard and Private Leaderboard are used referring to what is commonly understood as development or validation phase and test phase. This is important in the system development stage as it helps preventing systems from overfitting. We used 10-fold cross validation in the training setup.

As can be seen in Table \ref{tab:resultsleaderboard}, the ensemble system with additional data achieved the best performance.
This can be attributed to the use of large amounts of additional training data, a semi-supervised approach, and an ensemble model with many features.

\section{Conclusion and Future Work}

This paper presented an approach to categorizing patent applications in eight classes of the WIPO class taxonomy. Our system competed in the ALTA 2018 - Classifying Patent Applications shared task under the team name BMZ. Our best system is based on an ensemble of SVM classifiers trained on words and characters. It achieved 0.778 micro-averaged F1-Score and ranked first place in the competition among 14 teams.

We observed that expanding the training data using the WIPO datasets brought substantial performance improvement. This dataset is similar to that provided by the shared task organizers in terms of genre and topics and it contains 15 times more samples. The use of an ensemble-based approach prevented the system from overfitting and providing more robust predictions.

In future work we would like to use hierarchical approaches to classify patent applications using a more fine-grained taxonomy. Finally, we would also like to investigate the performance of deep learning methods 
for this task. 

\section*{Acknowledgments}

We would like to thank the ALTA 2018 shared task organizers for organizing this interesting shared task and for replying promptly to our inquiries. 

\bibliography{acl2018}

\begin{thebibliography}{}
\expandafter\ifx\csname natexlab\endcsname\relax\def\natexlab#1{#1}\fi

\bibitem[{Aletras et~al.(2016)Aletras, Tsarapatsanis, Preo{\c{t}}iuc-Pietro,
  and Lampos}]{aletras2016predicting}
Nikolaos Aletras, Dimitrios Tsarapatsanis, Daniel Preo{\c{t}}iuc-Pietro, and
  Vasileios Lampos. 2016.
\newblock {Predicting Judicial Decisions of the European Court of Human Rights:
  A Natural Language Processing Perspective}.
\newblock {\em PeerJ Computer Science\/} 2:e93.

\bibitem[{Benzineb and Guyot(2011)}]{benzineb2011automated}
Karim Benzineb and Jacques Guyot. 2011.
\newblock {Automated Patent Classification}.
\newblock In {\em Current challenges in patent information retrieval\/},
  Springer, pages 239--261.

\bibitem[{Brucker et~al.(2011)Brucker, Benites, and Sapozhnikova}]{BruckerBS11}
Florian Brucker, Fernando Benites, and Elena~P. Sapozhnikova. 2011.
\newblock {An Empirical Comparison of Flat and Hierarchical Performance
  Measures for Multi-Label Classification with Hierarchy Extraction}.
\newblock In {\em Procedings of KES Part {I}\/}.

\bibitem[{Cai and Hofmann(2007)}]{cai2007exploiting}
Lijuan Cai and Thomas Hofmann. 2007.
\newblock Exploiting known taxonomies in learning overlapping concepts.
\newblock In {\em IJCAI\/}. volume~7, pages 708--713.

\bibitem[{Fall et~al.(2003)Fall, T{\"o}rcsv{\'a}ri, Benzineb, and
  Karetka}]{fall2003automated}
Caspar~J Fall, Atilla T{\"o}rcsv{\'a}ri, Karim Benzineb, and Gabor Karetka.
  2003.
\newblock {Automated Categorization in the International Patent
  Classification}.
\newblock In {\em Acm Sigir Forum\/}. ACM, volume~37, pages 10--25.

\bibitem[{Farzindar and Lapalme(2004)}]{farzindar2004legal}
Atefeh Farzindar and Guy Lapalme. 2004.
\newblock {Legal Text Summarization by Exploration of the Thematic Structures
  and Argumentative Roles}.
\newblock {\em Proceedings of the Text Summarization Branches Out Workshop\/} .

\bibitem[{Gebre et~al.(2013)Gebre, Zampieri, Wittenburg, and
  Heskes}]{gebre2013}
Binyam~Gebrekidan Gebre, Marcos Zampieri, Peter Wittenburg, and Tom Heskes.
  2013.
\newblock {Improving Native Language Identification with TF-IDF Weighting}.
\newblock In {\em Proceedings of the BEA Workshop\/}.

\bibitem[{Grawe et~al.(2017)Grawe, Martins, and Bonfante}]{8260665}
M.~F. Grawe, C.~A. Martins, and A.~G. Bonfante. 2017.
\newblock {Automated Patent Classification Using Word Embedding}.
\newblock In {\em 2017 16th IEEE International Conference on Machine Learning
  and Applications (ICMLA)\/}. pages 408--411.

\bibitem[{Hofmann et~al.(2003)Hofmann, Cai, and
  Ciaramita}]{hofmann2003learning}
Thomas Hofmann, Lijuan Cai, and Massimiliano Ciaramita. 2003.
\newblock Learning with taxonomies: Classifying documents and words.
\newblock In {\em NIPS workshop on syntax, semantics, and statistics\/}.

\bibitem[{Jauhiainen et~al.(2018)Jauhiainen, Jauhiainen, and
  Lind{\'e}n}]{jauhiainen2018heli}
Tommi Jauhiainen, Heidi Jauhiainen, and Krister Lind{\'e}n. 2018.
\newblock Heli-based experiments in swiss german dialect identification.
\newblock In {\em Proceedings of the VarDial Workshop\/}.

\bibitem[{Katz et~al.(2014)Katz, II, and Blackman}]{Katz14}
Daniel~Martin Katz, Michael J.~Bommarito II, and Josh Blackman. 2014.
\newblock Predicting the behavior of the supreme court of the united states:
  {A} general approach.
\newblock {\em CoRR\/} abs/1407.6333.

\bibitem[{Malmasi et~al.(2016{\natexlab{a}})Malmasi, Dras, and
  Zampieri}]{malmasi2016ltg}
Shervin Malmasi, Mark Dras, and Marcos Zampieri. 2016{\natexlab{a}}.
\newblock {LTG at SemEval-2016 task 11: Complex Word Identification with
  Classifier Ensembles}.
\newblock In {\em Proceedings of SemEval\/}.

\bibitem[{Malmasi and Zampieri(2017)}]{malmasi-zampieri:2017:VarDial1}
Shervin Malmasi and Marcos Zampieri. 2017.
\newblock {German} dialect identification in interview transcriptions.
\newblock In {\em Proceedings of the VarDial Workshop\/}.

\bibitem[{Malmasi and Zampieri(2018)}]{malmasi2018challenges}
Shervin Malmasi and Marcos Zampieri. 2018.
\newblock {Challenges in Discriminating Profanity from Hate Speech}.
\newblock {\em Journal of Experimental \& Theoretical Artificial
  Intelligence\/} 30(2):187--202.

\bibitem[{Malmasi et~al.(2016{\natexlab{b}})Malmasi, Zampieri, and
  Dras}]{malmasi2016predicting}
Shervin Malmasi, Marcos Zampieri, and Mark Dras. 2016{\natexlab{b}}.
\newblock {Predicting Post Severity in Mental Health Forums}.
\newblock In {\em Proceedings of CLPsych Workshop\/}.

\bibitem[{Medvedeva et~al.(2018)Medvedeva, Vols, and
  Wieling}]{medvedeva2018judicial}
Masha Medvedeva, Michel Vols, and Martijn Wieling. 2018.
\newblock {Judicial Decisions of the European Court of Human Rights: Looking
  into the Crystal Ball}.
\newblock {\em Proceedings of the Conference on Empirical Legal Studies\/} .

\bibitem[{Molla and Seneviratne(2018)}]{ALTA2018}
Diego Molla and Dilesha Seneviratne. 2018.
\newblock {Overview of the 2018 ALTA Shared Task: Classifying Patent
  Applications}.
\newblock In {\em Proceedings of ALTA\/}.

\bibitem[{Sulea et~al.(2017{\natexlab{a}})Sulea, Zampieri, Malmasi, Vela, Dinu,
  and van Genabith}]{sulea2017exploring}
Octavia-Maria Sulea, Marcos Zampieri, Shervin Malmasi, Mihaela Vela, Liviu~P
  Dinu, and Josef van Genabith. 2017{\natexlab{a}}.
\newblock Exploring the use of text classification in the legal domain.
\newblock {\em arXiv preprint arXiv:1710.09306\/} .

\bibitem[{Sulea et~al.(2017{\natexlab{b}})Sulea, Zampieri, Vela, and van
  Genabith}]{sulea2017predicting}
Octavia-Maria Sulea, Marcos Zampieri, Mihaela Vela, and Josef van Genabith.
  2017{\natexlab{b}}.
\newblock {Predicting the Law Area and Decisions of French Supreme Court
  Cases}.
\newblock In {\em Proceedings of RANLP\/}.

\bibitem[{Tikk and Bir{\'o}(2003)}]{tikk2003experiment}
Domonkos Tikk and Gy{\"o}rgy Bir{\'o}. 2003.
\newblock Experiment with a hierarchical text categorization method on the
  wipo-alpha patent collection.
\newblock In {\em Proceedings of ISUMA 2003\/}.

\bibitem[{Tsochantaridis et~al.(2004)Tsochantaridis, Hofmann, Joachims, and
  Altun}]{tsochantaridis2004support}
Ioannis Tsochantaridis, Thomas Hofmann, Thorsten Joachims, and Yasemin Altun.
  2004.
\newblock Support vector machine learning for interdependent and structured
  output spaces.
\newblock In {\em Proceedings of the ICML\/}.

\bibitem[{Wongchaisuwat et~al.(2016)Wongchaisuwat, Klabjan, and
  McGinnis}]{wongchaisuwat2016}
Papis Wongchaisuwat, Diego Klabjan, and John~O McGinnis. 2016.
\newblock {Predicting Litigation Likelihood and Time to Litigation for
  Patents}.
\newblock {\em arXiv preprint arXiv:1603.07394\/} .

\bibitem[{Zampieri et~al.(2018)Zampieri, Malmasi, Nakov, Ali, Shon, Glass,
  Scherrer, Samard{\v{z}}i{\'c}, Ljube{\v{s}}i{\'c}, Tiedemann
  et~al.}]{zampieri2018language}
Marcos Zampieri, Shervin Malmasi, Preslav Nakov, Ahmed Ali, Suwon Shon, James
  Glass, Yves Scherrer, Tanja Samard{\v{z}}i{\'c}, Nikola Ljube{\v{s}}i{\'c},
  J{\"o}rg Tiedemann, et~al. 2018.
\newblock {Language Identification and Morphosyntactic Tagging: The Second
  VarDial Evaluation Campaign}.
\newblock In {\em Proceedings of VarDial Workshop\/}.

\bibitem[{Zhong et~al.(2018)Zhong, Xiao, Guo, Tu, Liu, Sun, Feng, Han, Hu, Wang
  et~al.}]{zhong2018overview}
Haoxi Zhong, Chaojun Xiao, Zhipeng Guo, Cunchao Tu, Zhiyuan Liu, Maosong Sun,
  Yansong Feng, Xianpei Han, Zhen Hu, Heng Wang, et~al. 2018.
\newblock {Overview of CAIL2018: Legal Judgment Prediction Competition}.
\newblock {\em arXiv preprint arXiv:1810.05851\/} .

\end{thebibliography}
\bibliographystyle{acl_natbib}

\end{document}